%% file: acl2019.tex
\documentclass[11pt,a4paper]{article}
\usepackage[hyperref]{acl2019}
\usepackage{times}
\usepackage{latexsym}

\usepackage{url}

\aclfinalcopy 
\def\aclpaperid{541} 

\usepackage{xcolor}
\usepackage{amsmath}
\usepackage{multirow}
\usepackage{graphicx}
\usepackage{booktabs}
\usepackage{enumitem}
\usepackage{xcolor}
\usepackage{array}
\newcolumntype{P}[1]{>{\centering\arraybackslash}p{#1}}
\newcommand\tab[1][0.5cm]{\hspace*{#1}}

\usepackage{xcolor}
\definecolor{accent}{HTML}{000000}
\definecolor{myblue}{rgb}{0,0.1,0.6}

\usepackage{amssymb}
\def\R{\mathbb{R}}

\usepackage{longtable}
\usepackage{pdflscape}
\usepackage{supertabular}

\title{Modeling financial analysts' decision making via the pragmatics and semantics of earnings calls} 

\author{Katherine A. Keith\\
College of Information and Computer Sciences \\
  University of Massachusetts Amherst \\
  {\tt kkeith@cs.umass.edu} \\\And
  Amanda Stent \\
  Bloomberg LP\\
  {\tt astent@bloomberg.net} \\}

\date{}

\begin{document}

\maketitle

\begin{abstract}
  Every fiscal quarter, companies hold
  \emph{earnings calls} in which company executives respond to
  questions from analysts. After these calls, analysts often change their
  \emph{price target recommendations}, which are used in equity research
  reports to help investors make decisions.  In this paper, we examine
  analysts' decision making behavior as it pertains to the language
  content of earnings calls.  We identify a set of 20 pragmatic features of  analysts' questions which we 
  correlate with analysts' pre-call investor recommendations. We also analyze the degree to which semantic and
  pragmatic features from an earnings call complement market data in predicting analysts' post-call changes in price
  targets.  Our results show that earnings calls are moderately predictive of
  analysts' decisions even though these decisions are influenced by a
  number of other factors including private communication with company
  executives and market conditions. A breakdown of
  model errors indicates disparate performance on
  calls from different market sectors.
\end{abstract}

\section{Introduction}

Financial analysts are key sell-side players in finance who are
employed to analyze, interpret, and disseminate financial information
\cite{brown2015inside}.  For the firms they cover, financial analysts
regularly release \emph{recommendations} to buy, hold, or sell the
company's stock, and stock \emph{price targets}.  Financial analysts'
forecasts are of value to investors \cite{givoly1980financial} and may
be better surrogates for market expectations than forecasts generated
by time-series models \cite{fried1982financial}.

\begin{table*}[t]
\small
  \centering
      \begin{tabular}{p{15cm}}
      \toprule 
      \textbf{Brian Nowak, Analyst: }
      \textbf{\color{purple}Thanks} for taking my questions. One on YouTube, \textbf{\color{purple}I guess}. Could you \textbf{\color{purple} just} talk to some of the qualitative drivers that are really bringing more advertising dollars on to \textbf{ \color{orange}YouTube}\textbf{\color{blue}?} And then I think \textbf{\color{green} last quarter} you had mentioned the \textbf{ \color{orange}top 100 advertiser} spending was \textbf{\color{orange} up 60\%} year-on-year on \textbf{\color{orange}YouTube}, wondering, if you could update us on that\textbf{\color{blue}?} And the second one on search, it sounds like mobile is accelerating. Where are you \textbf{\color{green}now} in the mobile versus desktop monetization gap\textbf{\color{blue}?} And, Sundar, how do you think about that \textbf{\color{green}long-term}\textbf{\color{blue}?} Do you see mobile being higher, reaching equilibrium\textbf{\color{blue}?} How do you see that trending\textbf{\color{blue}?} \\
      \midrule
      \textbf{Sundar Pichai, CEO: }
      On the \textbf{\color{orange}YouTube} one. \textbf{ \color{purple}Look, I mean,} the shift to video is a profound medium shift and especially in the context of mobile, \textbf{\color{purple}you know} and obviously users are following that. You're seeing it in \textbf{\color{orange}YouTube} as well as elsewhere in mobile. And so, advertisers are being increasingly conscious. They're being \textbf{\color{purple}very, very} responsive. So, we're seeing great traction there and we'll continue to see that. They are moving more off their traditional budgets to \textbf{\color{orange}YouTube} and that's where we are getting traction. On mobile search, to me, increasingly we see we already announced that \textbf{\color{orange}over 50\%} of our searches are on mobile. Mobile gives us very unique opportunities in terms of better understanding users and over time, as we use things like machine learning, \textbf{ \color{purple}I think} we can make great strides. So, my \textbf{ \color{green}long-term view} on this is, it is as-compelling or in fact even better than desktop, but it will take us time to get there. We're going to be focused till we get there. \\
      \bottomrule
  \end{tabular}
  \caption*{\label{t:ex}  Figure 1: Earnings calls are extremely complex examples of naturally-occurring discourse. In this example question-answer pair from a Google earnings call on October 27, 2016, the analyst asks \textbf{\color{blue} six distinct questions} in a single turn. Because the interaction originates as speech, there are \textbf{\color{purple}discourse markers and hedging.} The analyst and executive discuss \textbf{\color{orange}concrete entities and performance statistics} and \textbf{\color{green} past, present and future} performance.}
  \end{table*}
  
  Analysts' decisions are
  influenced by market conditions and private communications\footnote{\citet{brown2015inside}
    find over half of the 365 analysts they surveyed have five or more
    direct contacts per year with the CEO or CFO of companies they
    follow.}, so it is impossible to exactly
  reconstruct their decision making process. However, signals of
  analysts' decision making may be obtained by analyzing \emph{earnings
    calls}---quarterly live conference calls in which company
  executives present prepared remarks (the \emph{presentation}
  section) and then selected financial analysts ask questions (the
  \emph{question-answer} section).  Previous work has shown that
  earnings calls disclose more information than company filings alone
  \cite{frankel1999empirical} and influence investor sentiment in the
  short term \cite{bowen2002conference}. However, recently company executives and investors have questioned their value
\cite{marketwatch17,cnbc18}. 

Earnings calls are extremely complex, naturally-occurring examples of
discourse that are interesting to study from the perspective of
computational linguistics (see Figure~1).  In this work, we
examine analysts' decision making in the context of earnings calls in two
ways:
\begin{itemize}[noitemsep,leftmargin=*]
\item \textbf{Correlating analysts' question pragmatics with their pre-call judgements:} With domain experts, we  select a set of 20 pragmatic and discourse features which we extract from the questions of earnings calls. Then we correlate these with analysts' pre-call judgments and find \emph{bullish} analysts tend to be called on earlier in calls, and ask questions that are more positive, more concrete, and less about the past (\S\ref{s: dm1}). 
\item \textbf{Predicting changes in analysts' post-call forecasts:} 
 We use the pragmatic features, along with representations of the semantic content of earnings calls, to predict changes in analysts' post-call price targets. Since analysts have a deep understanding of market factors influencing a company's performance and have access to private information, our null hypothesis is that earnings calls are not predictive of forecast changes. However, our best model gives a reduction of $25\%$ in relative accuracy error over a majority class baseline (twice the reduction of a model using market data alone), suggesting there is signal in the noise. We also conduct pairwise comparisons of modeling features including: semantic vs.~pragmatic features, Q\&A-only vs.~whole-call data, and whole-document vs.~turn-level models (\S\ref{s:dm2}).
\end{itemize}

\section{Related work}\label{s:related-work}

NLP is used extensively for financial applications \cite{tetlock2008more,kogan2009predicting,leidner2010hunting,loughran2011liability,wang2013financial,ding2014using,peng2016leverage,li2017learning,rekabsaz2017volatility}. 
Earnings calls, in particular, are shown to be predictive of
investor sentiment in the short-term, including of increased stock
volatility and trading volume levels \cite{frankel1999empirical},
decreased forecast error and forecast dispersion
\cite{bowen2002conference}, and increased absolute returns for
intra-day trading \cite{cohen2012casting}. Although most prior work on
earnings calls treat each call as a single document,
\citet{matsumoto2011makes} find that the \emph{question-answer} portion of
the earnings call is more informative (in terms of intra-day absolute returns) than the \emph{presentation}
portion, and \citet{cohen2012casting} show firms ``cast'' earnings calls by disproportionately calling on bullish analysts.



Most prior applications of NLP to earnings calls use only
shallow linguistic features and correlation analyses, specifically
correlations between political bigrams and stock return volatility
\cite{hassan2016aggregate}; contrastive words and share prices
\cite{palmon2016does}; and euphemisms and earnings surprise
\cite{suslava2017stiff}.  Other work analyzes earnings calls from
a sociolinguistic perspective, including in terms of discourse
connectives \cite{camiciottoli2010discourse}, indirect requests
\cite{camiciottoli2009just}, unanswered questions
\cite{hollander2010does}, persuasion \cite{crawford2018persuasion}
and deception \cite{larcker2011detecting}. Focusing on only the audio
of earnings calls, \citet{mayew2012power} extract managers'
affective states using commercial speech software. In the work most
similar to ours, \citet{wang2014semiparametric} use named entities,
part-of-speech tags, and probabilistic frame-semantic features in
addition to unigrams and bigrams to correlate earnings calls with
financial risk, which they defined as the volatility of stock prices
in the week following the earnings call.


NLP-based corpus analyses of decision making are
rare. \citet{bevnuvs2014entrainment} analyze the impact of
entrainment on Supreme Court justices' subsequent decisions. Multiple
groups have examined the impact of various semantic and pragmatic
features on modeling opinion change using reddit ChangeMyView
discussions
(e.g.~\cite{hidey2017analyzing,jo2018attentive,musi2018did}), and
there has been other work on opinion change using other web discussion
data
(e.g.~\cite{tan2016winning,habernal2016argument,lukin2017argument}). Because
many factors influence decision making behavior, the fact that any
signal can be obtained from linguistic analyses of isolated language artifacts is scientifically interesting.

\begin{table}[t]
\small
  \centering
      \begin{tabular}{l r }
      \toprule
      Earnings calls total (2010-2017) & 12,285 \\
      \tab Train (2010-2015) & 9,770  \\
      \tab Validation  (2016) & 1,066 \\ 
      \tab Test (2017) & 1,449\\
      \midrule
      Unique companies & 642\\ 
      Total Q\&A sets & 573,550 \\
      Ave. Q\&A sets per doc. & 44.3 \\
      One call, ave. unique analysts speaking & 10.9 \\ 
      One call, ave. analysts w/ price targets & 9.6 \\
      Ave. num. of tokens per doc. & 8,761 \\
      Ave. turn length (num. tokens), Q\&A &  62.7 \\
      \bottomrule
  \end{tabular}
  \caption{Data statistics for S\&P 500 companies' earnings calls.  A Q\&A set consists of two or more turns, one containing an analyst's question(s) and the rest containing company representatives' answer(s). 
  \label{t:data}}
\end{table}

\section{Data and pre-processing}
Our data\footnote{In Appendix A in supplemental material we provide the stock tickers for the calls in our data; the corpus can be re-assembled from multiple sources, such as \url{https://seekingalpha.com/}.} consists of transcripts of 12,285 earnings calls held between January 1, 2010 and December 31, 2017. In order to control for analyst coverage effects (larger companies with a greater market share will typically be covered by more analysts), 
we include only calls from S\&P 500 companies. We split the data by year into training, validation and testing sets (see Table \ref{t:data}).


The transcripts are XML files with metadata specifying speaker turn boundaries and the name of the speaker (or ``Operator'' for the call operator). In order to identify \emph{speaker type} (analyst or company representative) we use the following heuristic: if the transcript explicitly includes the speaker type with the speaker name (e.g.~``John Doe, Analyst"), we do exact string matching for ``, Analyst"; else, we assume the names of speakers between the first and second operator turns (i.e. in the \emph{presentation} section) are those of company representatives and all other speakers are analysts. We manually checked this heuristic on a few dozen documents and found it to have high precision. 

We remove turns spoken by the operator as well as turns that have fewer than 10 tokens since manual analysis revealed the latter were largely acknowledgment and greeting turns (e.g.~``Thank you for your time" and ``You're welcome"). We also lexicalized named entities and represented them as a single token. We obtained tokenization, part of speech tagging, and dependency parsing via a proprietary NLP library\footnote{Bloomberg's \texttt{libnlp}}.

\begin{table*}[t]
  \centering
 \small 
      \begin{tabular}{l l l l r}
 \toprule
      \textbf{No.} & \textbf{Pragmatic Lexicon} & \textbf{Examples} & \textbf{Source} & \textbf{Num. terms} \\
      \toprule 
      10 &Positive sentiment, financial & booming, efficient, outperform  & LM & 354 \\
      10 & Positive sentiment, general-purpose & perfection, enthrall, phenomenal & T & 2,507 \\
      11 & Negative sentiment, financial & accidents, recession, stagnant  & LM & 2,353 \\
      11 & Negative sentiment, general-purpose & cheater, devastate, loathsome & T & 3,692 \\
       12 & Hedging, unigrams &  basically, generally, sometimes & PH  & 79 \\
     12 & Hedging, multi-word & a little, kind of, more or less & PH & 39 \\
      13 & Weak Modal & appears, could, possibly & LM & 27 \\
      13 & Moderate Modal & likely, probably, usually & LM & 14 \\
      13 & Strong Modal & always, clearly, undoubtedly & LM & 19 \\
      14 & Uncertain & assume, deviate, turbulence & LM & 297 \\
      15 & Constraining & bounded, earmark, indebted & LM & 184 \\
      16 & Litigious & adjudicate, breach, felony, lawful & LM & 903 \\
      \bottomrule 
  \end{tabular}
  \caption{\label{t:lexicon} Detailed examples and the number of words for lexicons used as pragmatic features. LM is  \protect\cite{loughran2011liability}, PH is  \protect\cite{prokofieva2014hedging}  and T is \protect\cite{taboada2011lexicon}. Feature numbers (No.) correspond to the text description in \S\ref{ss:prag}.}
\end{table*}

\begin{table*}[t]
\small
  \centering
      \begin{tabular}{l l >{\raggedright}p{10cm} r}
       \toprule
      \textbf{No.} & \textbf{Pragmatic Feat.}  & \textbf{Example}  & \textbf{Score}  \\
      \toprule
      6 &  Concreteness & Yes. \colorbox{yellow}{Andrew} for \colorbox{yellow}{the\_quarter} the total inter-company sales for \colorbox{yellow}{the\_first\_quarter} was roughly \colorbox{yellow}{4.6\_million} and about \colorbox{yellow}{600,000} was related to medical, it was \colorbox{yellow}{4\_million} via \colorbox{yellow}{DSS}. 
      & 0.29 \\
      \midrule
      10 & Positive sentiment  & \colorbox{yellow}{Good} morning, gentlemen. \colorbox{yellow}{Nice} job on the \colorbox{yellow}{rebound} quarter.  
      & 0.33 \\
      \midrule
     11 & Negative sentiment & And this is a \colorbox{yellow}{slightly} delicate question. With some of the \colorbox{yellow}{terrible} events that have been happening, what is this \colorbox{yellow}{duty} or potential \colorbox{yellow}{liability} or \colorbox{yellow}{cost} of insurance? 
      & 0.15 \\
      \midrule
     12 &  Hedging & It \colorbox{yellow}{may} vary Michael. So, \colorbox{yellow}{some} \colorbox{yellow}{might} be \colorbox{yellow}{much} better than that, but then you got \colorbox{yellow}{some} of that -- that's not as \colorbox{yellow}{much} right. So, all-in, yeah. 
      & 0.22\\
      \bottomrule 
  \end{tabular}
  \caption{Pragmatic features as highlighted tokens. Note, named entities are lexicalized (e.g.~``4.6\_million"). 
  Feature numbers (No.) correspond to the text description in \S\ref{ss:prag}.
  \label{t:prag}
  }
\end{table*}

\section{Pragmatic correlations with analysts' pre-call judgments} \label{s: dm1}
We are interested in whether and how the forms of analysts' questions reflect their pre-call judgments about companies they cover. Analysts' questions are complex: a single turn may contain several questions (or answers). An example question-answer pair is shown in Figure~1.

We compute Pearson correlations between linguistic features indicating certainty, deception, emotion and outlook (\S\ref{ss:prag}) and the \emph{type} of analyst (bullish, bearish, or neutral) asking the question. 
We use a mapping of analysts' recommendations to a 1-5 scale\footnote{Qualitative analyst rating labels vary from firm to firm. For example, some firms use the standard ``buy, hold, sell" labels while others might use different labels such as ``outperform, peer perform, underperform." We use ratings  from a proprietary financial database that have been manually normalized to 1-5 scale.} where a 1 denotes ``strong
sell" and a 5 denotes ``strong buy." We label each analyst according to
their recommendation of the company before the earnings call:
\begin{itemize}[noitemsep]
\itemsep0em
\item \emph{bearish} if analysts give a company a 1 or 2, 
\item \emph{neutral} if they give a 3, and 
\item \emph{bullish} if they give a 4-5.  
\end{itemize}
We
have analyst recommendations for 160,816 total question turns and the
distribution over analyst labels is 4.5\% bearish, 35.7\% neutral, and
59.7\% bullish.  
Following other correlation work in NLP
\cite{preoctiuc2015role,holgate2018swear}, we use Bonferroni
correction to address the multiple comparisons problem.

\subsection{Pragmatic lexical features} \label{ss:prag}
We extract 20 pragmatic features from each turn by gathering existing
hand-crafted, linguistic lexicons for these concepts\footnote{Appendix
  B in supplemental material gives details about the sources of our
  lexicons.}. See Table \ref{t:lexicon} for statistics about the
lexicons and Table~\ref{t:prag} for examples.

\textbf{Named entity counts and concreteness ratio.}
For each turn, we calculate the number of named entities in five coarse-grained groups constructed from the fine-grained entity types of OntoNotes\footnote{Version 5, \url{https://catalog.ldc.upenn.edu/docs/LDC2013T19/OntoNotes-Release-5.0.pdf} Section 2.6.} \cite{hovy2006ontonotes}: 
(1) events, 
(2) numbers, 
(3) organizations/locations, 
(4) persons, and 
(5) products.  
We also calculate (6) a \emph{concreteness ratio}: the number of named entities in the turn divided by the total number of tokens in the turn.

\textbf{Predicate-based temporal orientation.}
Temporal orientation is the emphasis individuals place on the past, present, or future. Previous work has shown correlations between ``future intent" extracted from query logs and financial market volume volatility \cite{hasanuzzaman2016collective}.  We determine the temporal orientation of every predicate in a turn. We extract OpenIE predicates via a re-implementation of PredPatt \cite{white2016universal}. 
 For each predicate, we look at its Penn Treebank part-of-speech tag and use a heuristic\footnote{If the part-of-speech tag for the predicate is \textsc{VBD} or \textsc{VBN} the temporal orientation is ``past"; otherwise if it is \textsc{VB, VBG, VBP}, or \textsc{VBZ} it is ``present" unless the predicate has a dependent of the form \emph{will, 'll, shall} or \emph{wo} indicating ``future'', \emph{are is, am,} or \emph{are} indicating ``present'', or \emph{was} or \emph{were} indicating ``past''.} to determine if it is ``past," "present," or ``future." :  We calculate the number of (7) ``past'' oriented predicates, (8) ``present'' oriented predicates and (9) ``future'' oriented predicates in each turn. 

\textbf{Sentiment.}
We calculate the ratio of (10) positive sentiment terms and (11) negative sentiment terms to the number of tokens in each turn. We use the financial sentiment lexicons developed by \citet{loughran2011liability} from fourteen years of 10-Ks. We supplement these with a general-purpose sentiment dictionary 
\cite{taboada2011lexicon}, to account for the relative informality of earnings calls.

\textbf{Hedging.}
We calculate (12) the ratio of hedges to tokens in each turn.
Hedges are lexical choices by which a speaker indicates a lack of commitment to the content of their speech \cite{prince1982hedging}. We use the single- and multi-word hedge lexicons from \cite{prokofieva2014hedging}.

\textbf{Other lexicon-based features.} 
We compute the ratios of (13) modal, (14) uncertain, (15) constraining, and (16) litigious terms in each turn using the respective lexicons from \citet{loughran2011liability}. In each case, we compute the ratio of terms in the category to the number of tokens in the turn. 

\textbf{Other pragmatic features.}
We also calculate (17) the turn order, (18) the number of tokens, (19) the number of predicates, and (20) the number of sentences in each turn. 

\begin{table}[t]
  \centering
  \small
      \begin{tabular}{l l r r}
\toprule 
      No. & Feature & Pearson's \emph{r} & p-value \\
      \toprule
1 &Named entities event&$0.0041$&$0.0999$\\
2 & Named entities number&$0.0064$&$0.0099$\\
3*& Named entities org. &$0.0185$&$<1\mathrm{e}^{-4}$\\
4*& Named entities person &$0.0247$&$<1\mathrm{e}^{-4}$\\
5& Named entities product&$0.0022$&$0.3777$\\
6* & Concreteness ratio &$0.0115$& $<1\mathrm{e}^{-4}$\\
7* & Num past preds &$-0.0086$&$0.0006$\\
8 & Num present preds &$0.0052$&$0.0378$\\
9 & Num future preds &$0.0033$&$0.1914$\\
10* & Sentiment positive&$0.0162$&$<1\mathrm{e}^{-4}$\\
11* & Sentiment negative&${-0.0104}$&$<1\mathrm{e}^{-4}$\\
12 & Hedging&$0.0017$&$0.5019$\\
13 & Modal&$0.0075$&$0.0028$\\
14 & Uncertainty&$0.0055$&$0.0287$\\
15 & Constraining&$0.0005$&$0.8399$\\
16 & Litigiousness&$-0.0072$&$0.0037$\\
17* & Turn order&$-0.1034$&$<1\mathrm{e}^{-4}$\\
18 & Num. tokens & $0.0050$ & $0.0459$ \\
19 & Num predicates&$0.0011$&$0.6692$\\
20 & Num sents. & $0.0043$ & $0.0854$\\
      \bottomrule  
  \end{tabular}
  \caption{Results from Pearson correlations of pragmatic lexical features from \S\ref{ss:prag} and prior-to-call labels of analysts, ( \emph{bearish}, \emph{neutral}, or \emph{bullish}). Statistical significance after Bonferroni correction is marked by (*) for $p < 0.0025$. Total 160,816 question turns. 
  \label{t:results1}
  }
\end{table}

\subsection{Interpretation of correlation results.} \label{s:corr-results}
Full results for the pragmatic correlation analysis are given in Table \ref{t:results1}. For a number of features the correlations are not statistically significant. However, we expand upon the statistically significant results for negative ($-$) and positive ($+$) correlations with the bullishness of an analyst:
\begin{itemize}[leftmargin=*]
\itemsep0em
\item \emph{($+$) Bullishness and turn order.} This suggests bullish analysts tend to be called on earlier in the call and bearish and neutral analysts tend to be called on later in the call which confirms the conclusion of \citet{cohen2012casting}. 

\item \emph{($+$) Bullishness and positive sentiment.} Bullish analysts tend to ask more positive (less negative) questions and the reverse is true for neutral/bearish analysts. Intuitively, this makes sense since bullish analysts are more favorable towards the firm and thus probably cast the firm in a positive light.

\item \emph{($+$) Bullishness and entities.} Here we find that bullish analysts are slightly more concrete in their questions towards the company and tend to ask more about organizations and people. 

\item \emph{($-$) Bullishness and past predicates.} This suggests bearish and neutral analysts tend to talk about the past more.  
\end{itemize}

These correlations could be used by journalists and investors to flag questions that follow atypical patterns for a particular analyst.

\section{Predicting changes in analysts' post-call forecasts}
  \label{s:dm2}

We are interested in what earnings-call related information is indicative of analysts' subsequent decisions to change (or not change) their \emph{price target} after an earnings call. 
A \emph{price target} is a projected future price level of asset; for example, an analyst may give a stock that is currently trading at \$50 a six-month price-target of \$90 if they believe the stock will perform better in the future.

We design experiments to answer the following \textbf{research questions}: (1) Is the text of earnings calls predictive of analysts' changes in price targets from before to after the call? This is an open research question since analysts may change their price targets at any time and consider external information (e.g.~current events or private conversations with company executives); (2) If the text is predictive, is the text more predictive than market-based features such as the company's stock price, volatility, and earnings? (3) If the text is predictive, what linguistic aspects (e.g.~pragmatic vs.~semantic) are more predictive and with which feature representations? (4) Is the \emph{question-answer} portion of the call more predictive than the \emph{presentation} portion? (5) Does a turn-based model of the call provide more signal than ``single document'' representations?

\subsection{Representing analysts' forecast changes}

We model analysts' changes in forecasts as both a regression task and a 3-class classification task because different formulations may be of interest to various stakeholders\footnote{For instance, investors may care more about small changes in forecast price targets whereas journalists may care more about relative changes (e.g.~ whether an earnings call will move analysts' price targets up or down).}.


\begin{figure}[t]
\centering
\includegraphics[width=\columnwidth]{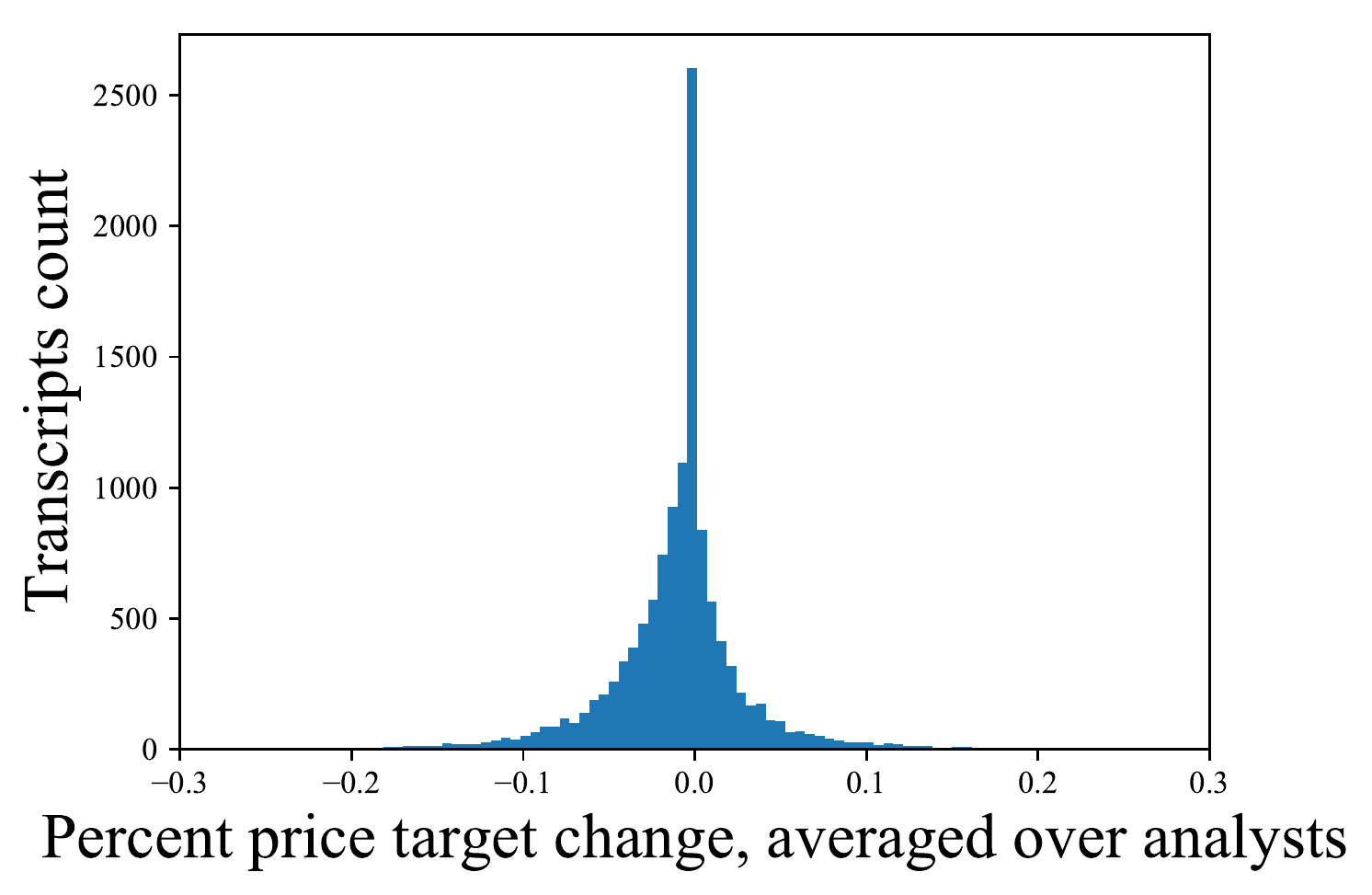}
\caption{Distribution across the entire corpus of prediction y-values, percentage price change in analyst price targets.  
\label{f:prices}}
\end{figure}

\textbf{Regression.}
For each earnings call in our corpus, $i \in \mathcal{D}$, and each analyst in the set of analysts covering that call,  $j \in J_i$, let $b_j$ be the price target of analyst $j$ before the call and let $a_j$ be the price target after the call\footnote{Because the company holding the earnings call chooses which analysts to call on for questions, our data includes analyst ratings and recommendations from analysts who do \emph{not} ask a question in a call. Also, because individual analysts' recommendations may be sold to different vendors, we do not have analyst ratings and recommendations for all analysts who ask questions in our data.}. Then the \emph{average percent change in analysts' price targets} is 
\begin{equation}
y_i = \frac{1}{|J_i|} \sum_{j \in J_i} \frac{a_j - b_j}{b_j}.
\end{equation}
See Figure \ref{f:prices} for the distribution of $y_i$. 

\noindent
Since analysts can report price targets at any time, we set cut-off points for $a_j$ and $b_j$ to be 3 months before and 14 days after the earnings call date respectively (a majority of analysts who change their price targets do so within two weeks after a call). 

\begin{table}[t]
\small
  \centering
      \begin{tabular}{lrrrr}
      \toprule
      Dataset & ${-1} $ & ${0}$ & ${1}$ \\
      \toprule
      Train & 33.3\% & 38.3\% & 28.4\% \\
      Validation & 29.2\% & 30.5\% & 40.3\%\\
      Test & 33.6\% & 38.7\% & 27.7\% \\
      \bottomrule
  \end{tabular}
  \caption{Percentage of examples in each class (${-1}, 0, 1$) for the training, validation, and test sets. 
  \label{t:class-per}}
\end{table}

\textbf{Classification.}
We create three (roughly equal) classes (\emph{negative, neutral,} and \emph{positive} change) by binning the $y_i$ values calculated in the equation above into thirds. For each earnings call $i$, $c_i = {-1}$ if $y_i < -0.0167$, $c_i = 0$ if $-0.0167 \le y_i \le 0.0$,  and $c_i = 1$ if $0 < y_i$. Table~\ref{t:class-per} shows the class breakdown for each split of the data.

\subsection{Features}
We compare models with market-based, pragmatic, and semantic features.

\subsubsection{Market features}
For each company and call in our dataset, we obtain 10 market features for the trading day prior to the call date: open price, high price, low price, volume of shares, 30-day volatility, 10-day volatility, price/earnings ratio, 
relative price/earnings ratio, EBIT yield, and earnings yield\footnote{See Appendix B in supplemental material for detailed definitions of these finance terms.}.
We impute missing values for these features using the mean value of features in the training data\footnote{There are missing values for less than $1\%$ of the data. The missing values are mainly due to company acquisitions and changing of company names.}. We scale features to have zero mean and unit variance. 

\setlength{\tabcolsep}{5pt}

\begin{table*}[t]
  \centering
  \small
      \begin{tabular}{l l l  r r r l r r r}
      \toprule
      & & \multicolumn{4}{c}{Regression Task} & \multicolumn{4}{c}{Classification Task} \\
      \cmidrule(lr){3-6} \cmidrule(lr){7-10}
     Feature type &  Feature & Model & MSE & $R^2$ & \% err. & Model & Acc. & F1 & \% err. \\
      \toprule
      Baselines & Random (ave. 10 seeds)  & -- & $0.32987$ & $-199.9$ & --
      & -- & $0.340$ & $0.338$ & -- \\
     & Training mean & -- & $0.00165$ & $-1\mathrm{e}^{-5}$ & 0.0 
     & -- & -- & -- & -- \\
     &  Predict 0 & -- & $0.00177$ & ${-0.072}$ & -- 
     & -- & -- & -- & --\\
     & Predict majority class & -- & -- & -- & -- & -- & $0.387$ & $0.186$ & 0.0 \\
     \midrule
       Market & Market  
       & RR & $0.00160$ & $0.0478$ & $3.0$ 
      & LR & $0.435$ & $0.408$ & $12.4$ 
       \\
        \midrule 
     Semantic 
     & Bag-of-words
     & RR-WD & $0.00140$ & $0.1500$ & $15.2$ 
     & LR-WD & $\textbf{0.482}$ & $\textbf{0.475}$ & $\textbf{24.8}$
     \\
     && RR-Q\&A & $0.00165$ & $-0.0043$ & $0.0$
     & LR-Q\&A & $0.388$ & $0.189$ & $0.3$ \\
      \cmidrule(lr){2-10}
     & doc2vec
      & RR-WD & $\textbf{0.00137}$ & $\textbf{0.1718}$ & $\textbf{17.0}$
      & LR-WD & $0.479$ & $0.468$ & $23.8$  
      \\
      && RR-Q\&A & $0.00165$ & $-0.0031$ & $0.0$
     & LR-Q\&A & $0.385$ &  $0.220$ & $0.5$ \\
     & & LSTM & $0.00155$ & $0.0598$ & $6.1$  
     & LSTM &  $0.442$ & $0.400$ & $14.2$ \\
     \midrule 
     Pragmatic & Pragmatic lexicons
       & LSTM & $0.00164$ & ${-0.0020}$ & $0.6$ 
       & LSTM & $0.415$ & $0.368$ & $7.2$
       \\  
     \midrule
     Fusion & doc2vec + prag & LSTM & $0.00155$ & $0.0573$ & $6.1$ 
     & LSTM & $0.461$ & $0.460$ & $19.1$ \\
     \midrule 
     Ensemble & doc2vec + prag + market & Ens. & $0.00154$ & $ 0.0619$ & $6.7$  
       & Ens. & $0.460$ & $0.461$ & $18.9$ 
        \\ 
     \bottomrule 
       \end{tabular}
  \caption{Test-set regression and classification results. Models are ridge regression (RR), long short-term memory networks (LSTM), logistic regression (LR), and ensemble (Ens.). \emph{WD} denotes whole-document models, while \emph{Q\&A} denotes Q\&A-only models. Evaluation metrics are mean squared error (MSE), the coefficient of determination ($R^2$), accuracy (Acc.), and macro-level F1. For regression, percent error reduction (\% err.) is from the MSE of the baseline of predicting the training mean; for classification, it is from the accuracy of predicting the majority class. 
  \label{t:results2}
  }
\end{table*}

\setlength{\tabcolsep}{6pt}

\subsubsection{Semantic features}

\textbf{Doc2Vec.}
We use the \emph{paragraph vector} algorithm proposed by \citep{le2014distributed} to obtain 300-dimensional document embeddings. Depending on the model, we train doc2vec embeddings over whole calls, question-answer sections only, and individual turns. Using the \texttt{Gensim}\footnote{Version 3.6.0} implementation \cite{rehurek_lrec}, we train \emph{doc2vec} models for 50 epochs and ignore words that occur less than 10 times in the respective training corpus. 

\textbf{Bag-of-words.} We lowercase tokens, augment them with their
parts of speech, and then limit the vocabulary to the top 100K content
words\footnote{UD Part of speech tags ADJ, ADV, ADV, AUX, INTJ, NOUN,
  PRON, PROPN, VERB.} in the training data. Depending on the model, we calculate bag-of-words feature vectors over the whole document, over the \emph{Q\&A} section, and over each turn separately.

\subsubsection{Pragmatic features} We combine the 20 pragmatic features described in Section \ref{ss:prag} into a single feature vector. These features are only used in our turn-level models.

\subsection{Models}
We use several models to predict changes in analysts' forecasts.

\subsubsection{Whole-document models}

\textbf{Ridge regression\footnote{We also tried Kernel ridge regression with a Gaussian (RBF) kernel, which gave similar results. See Appendix C for more details.}.}  
For regression, we use ridge regression\footnote{Implemented with \texttt{scikit-learn}.} which has a loss function that is the linear least squares function and is regularized with an L2-norm. 
To tune hyperparameters, we perform a five-fold cross-validation grid search over the regularization strength\footnote{$\alpha$ in scikit-learn for values  $10^{-3}$ to $10^8$ by logarithmic scale.}. We evaluate on mean squared error (MSE) and the coefficient of determination ($R^2$) scores.

\textbf{Logistic regression\footnote{We also tried support vector machines; see Appendix C.}.}
For classification, we train logistic regression with a L2 penalty\footnote{Implemented with \texttt{sklearn}.} and we tune $C$, the inverse regularization constant, via a grid search and 5-fold cross validation on the training set. We evaluate validation and test sets using accuracy and macro F1 scores. 

\subsubsection{Q\&A-only models}

In order to compare the relative influence of the \emph{presentation} versus \emph{question-answer} sections of the earnings calls, we remove the \emph{presentation} portion of each call and only predict on the $Q\&A$ portion\footnote{Of the 12,285 documents, there were 246 that only contained the \emph{presentation} section and did not have the \emph{question-and-answer} section. In the Q\&A modeling we completely remove these documents.
}. Except for this difference, Q\&A-only models are identical to whole-document models.

\subsubsection{Turn-by-turn models}

\textbf{LSTM for regression.}  We model transcripts as a sequence of
turns using long-short term memory networks (LSTMs)
\cite{hochreiter1997long}.  Let $x_t \in \R^k$ be the input vector at
time $t$ for embedding dimension $k$, and let $L$ be the total length
of the sequence. Each $x_t$ is fed into the LSTM in order and produces
a corresponding output vector $h_t$. Then the final output vector is
passed through a linear layer $y = w^y h_L + b^y$ for output $y \in
\R$ with $w^y \in R^k$.  For a given mini-batch $b$, $L_b$ is fixed as the maximum number of turns among all documents and the sequences for the other documents in the mini-batch are padded. The network is trained with mean squared
error (MSE) loss. 

\textbf{LSTM for classification.}
The LSTM architecture for classification is similar to that used for regression except that there is an additional softmax layer after the final linear layer. This network is trained with cross-entropy loss.

Both LSTMs are trained via a grid search over the following hyperparameters: learning rate, hidden dimension, batch size, number of layers, and L2-penalty (a.k.a. weight decay). 
The networks are written in Pytorch\footnote{\url{https://pytorch.org/}} and optimized with Adam \cite{kingma2014adam} . 
 
%

\subsubsection{Fusion and ensembling}

\textbf{Early fusion.} We use early fusion \cite{atrey2010multimodal} to combine semantic and pragmatic feature vectors at every turn and feed these into a LSTM.
 
 \textbf{Ensembling via stacking.} 
 We use ``stacked generalization" \cite{wolpert1992stacked} (a.k.a.~``stacking") to combine fusion and market-based models. For regression, we take the output values from the fusion and market-based models as features into a ridge regression model. For classification, we take the three-dimensional probability vector output from the fusion and market-based models and concatenate these as features into a logistic regression model. In both cases, hyperparameters are tuned on validation data.
 
\subsubsection{Baselines.} 
We compare against several baselines: (1) random, drawing a random variable from a Gaussian centered at the mean of the training data, (2) predicting the mean change in forecast across all documents in the training set (regression), and (3) predicting 0, the majority class (classification).  

\begin{figure}[t]
\centering
\includegraphics[width=\columnwidth]{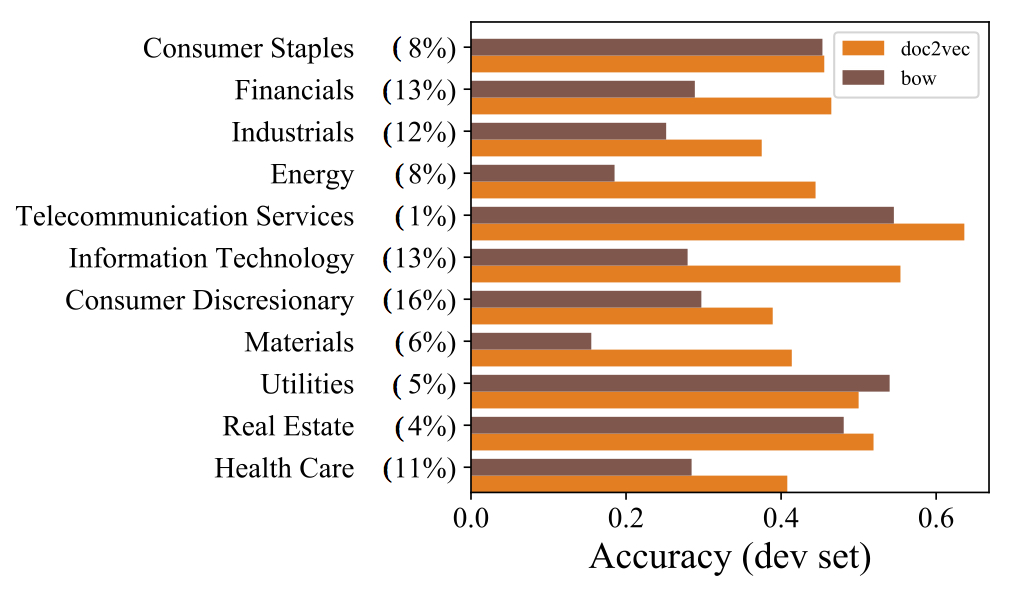}
\caption{Per-industry breakdown of errors on the validation set for \emph{doc2vec} (overall dev acc. 44.6\%) and \emph{bag-of-words} (bow) (overall dev acc. 30.4\%) models. Y-axis denotes the 11 GICS industries and their percentage of documents across the entire corpus. 
\label{f:industry}}
\end{figure}

\subsection{Results.} 
See Table \ref{t:results2} for full results. We address our original \textbf{research questions} from the beginning of \S\ref{s:dm2}. 

(1) \textbf{Predictiveness.} We find earnings calls are moderately predictive of changes in analysts' forecasts, with an almost $25\%$ relative error reduction in classification accuracy from the baseline of predicting the majority class. 
While the accuracy of our best model may seem modest, for this task, analysts' decisions can be influenced by many external factors outside of the text itself and our ability to find any signal among the noise may be interesting to financial experts. 

(2) \textbf{Text vs.~market.} Semantic features are more predictive of changes in analysts' price targets than market features (a 24.8\% error reduction over baseline for bag-of-words and a 23.8\% reduction for doc2vec, vs. a 12.4\% error reduction for market features).

 (3)  \textbf{Semantic vs.~pragmatic.} Semantic features (doc2vec and bag-of-words) are more predictive than pragmatic features. This suggests the semantic content of the earnings call is important in how analysts make decisions to change their price targets. 
 
 (4) \textbf{Q\&A-only vs.~whole-doc.} Contrary to
\citet{matsumoto2011makes} who find the \emph{question-answer}
portions of earnings calls to be most informative, we find the Q\&A-only
models are much less predictive for doc2vec (accuracy
0.479 vs.~0.385) and bag-of-words (accuracy 0.482 vs.~0.388) models.

(5) \textbf{Whole-doc vs turn-level.} Whole-document models are more predictive
than turn-level models (the best LSTM model achieves 19.1\% error reduction over baseline, vs. 24.8\% for the best whole-doc model). We hypothesize that
turn-level models might capture more signal if they incorporate speaker
metadata, e.g.~the role of the speaker or the analysts' pre-calls
judgment for the company.  Although whole-document models are more predictive,
 turn-level analyses of analysts' behavior may be more useful to alerting stakeholders to predictive signals in real-time (e.g.~an important analyst analyst question mid-way through a live earnings call) since financial markets can vary significantly in short time periods. 
 
\textbf{Breakdown of results by industry.}
We analyze errors on the validation data by segmenting earnings calls by each company's Global Industry Classification Standard (GICS) sector\footnote{See \url{https://www.msci.com/gics}. There are 11 broad industry sectors.}.
See Figure \ref{f:industry} for the breakdown results. Notably, the bag-of-words model performs almost 2.5 times worse on earnings calls from the \emph{Materials} sector versus the \emph{Utilities} and \emph{Telecommunication Services} sectors. This suggests industry-specific models may be important in future work. 

\section{Conclusions and future work} \label{s:conc}
In this work we (a) correlate pragmatic features of analysts'
questions with the pre-call judgment of the questioner, (b) explore
the influence of market, semantic and pragmatic features of earnings
calls on analysts' subsequent decisions.  We show that bullish
analysts are more likely to ask slightly more positive and concrete
questions, talk less about the past, and be called on earlier in a
call. We also demonstrate earnings calls are moderately predictive of
changes in analysts' forecasts.

Promising directions for future research include examining additional
features and feature representations: pragmatic features such as
formality \cite{pavlick2016empirical} or politeness
\cite{danescu2013computational}; acoustic-prosodic features from
earnings call audio; more sophisticated semantic representations such
as claims \cite{lim2016claimfinder}, automatically induced
entity-relation graphs \cite{bansal2017relnet} or question-answer
motifs \cite{zhang2017asking} (these representations are non-trivial
to construct because a single turn may contain many questions or
answers); or even discourse structures. The models used in this work
aim to be just complex enough to determine whether useful signals
exist for this task; future modeling work could include training a
complete end-to-end system such as a hierarchical attention network
\cite{yang2016hierarchical}, or building industry-specific models.


\section*{Acknowledgments}
We thank Sz-Rung Shiang, Christian Nikolay, Clay Elzroth, David Rosenberg, and Daniel Preotiuc-Pietro for guidance early on in this work. We also thank Abe Handler, members of the UMass NLP reading group, and anonymous reviewers for their valuable feedback. This work was partially supported by NSF IIS-1814955. 

\bibliography{acl2019}
\bibliographystyle{acl_natbib}

\appendix
\input{appendixA.tex}
\input{appendixB.tex}


\end{document}

%% file: appendixA.tex
\section{Calls Used in This Work}
See~\url{https://kakeith.github.io/attach/acl2019_supplement.pdf} for the list of earnings calls used in this work, i.e. all earnings call transcripts available to us for every company that was in the S\&P 500 on the date of the call, between 2010 and 2017 inclusive. The overall number of S\&P 500 companies in our data (642) is greater than 500 because we look at company inclusion in the S\&P 500 index \emph{daily}; companies regularly enter and leave this index. 

%% file: appendixB.tex
 \begin{table*}[h!]
  \centering
  \small
      \begin{tabular}{l l l  r r r l r r r}
      \toprule
      & & \multicolumn{4}{c}{Regression Task} & \multicolumn{4}{c}{Classification Task} \\
      \cmidrule(lr){3-6} \cmidrule(lr){7-10}
     Feature type &  Features & Models & MSE & $R^2$ & \% err. & Models & Acc. & F1 & \% err. \\
      \toprule
           Market & Market  
            & GK & $0.00163$ & $0.0117$ & $1.2$ 
       & SVM & $0.423$ & $0.379$ & $9.3$
       \\
     \midrule
     Semantic 
     & Bag-of-words
     & GK & $0.00152$ & $0.0765$ &$7.9$ 
     & SVM & -- & -- & --
     \\ 
      \cmidrule(lr){2-10}
     & doc2vec
      & GK & $0.00140$ & $0.1513$ & $15.2$ 
      & SVM & $0.476$ & $0.455$ & $23.0$ \\
         \bottomrule 
       \end{tabular}
  \caption{Results on the test set for additional models. Comparable to Table 6 in the main document.
  \label{t:results1}
  }
\end{table*}

\section{Additional Details Regarding Definitions and Sources of Features}

\subsection{Market features} 
The \textbf{relative price/earnings ratio} is a stock's price/earnings ratio relative to the price/earnings ratio of a relevant index, in this case the S\&P 500. 

\noindent The \textbf{EBIT yield} is equivalent to the (trailing 12-month operating income per share / last price) *100. 

\noindent The \textbf{earnings yield} is equivalent to the (trailing 12-month earnings per share before extraordinary items)/ last price) *100. 

\subsection{Pragmatic lexicons}

\subsubsection{OntoNotes five-coarse grained groups} 
For the pragmatic entity features, we construct five coarse-grained groups from the fine-grained entity types of OntoNotes\footnote{Version 5, {\small \url{https://catalog.ldc.upenn.edu/docs/LDC2013T19/OntoNotes-Release-5.0.pdf}} Section 2.6} \cite{hovy2006ontonotes}: 
(1) events (OntoNotes' \textsc{event}); 
(2) numbers (OntoNotes' \textsc{ordinal, money, percent, cardinal, time, date, quantity}); 
(3) organization/locations (OntoNotes' \textsc{loc, norp, facility, gpe, location, organization}); 
(4) persons (OntoNotes' \textsc{person}); and
(5) products (OntoNotes' \textsc{product}).  

\subsubsection{Sentiment}

As a financial sentiment lexicon, We used the positive and negative word lists from:\\
 {\small \url{https://sraf.nd.edu/textual-analysis/resources/}}\footnote{Archived at {\small \url{https://web.archive.org/web/20181203160914/https://sraf.nd.edu/textual-analysis/resources/}}}\\
 \cite{loughran2011liability}, as retrieved in July 2018. 

As a general sentiment lexicon, we used the SO-CAL dictionary from:\\ {\small \url{https://github.com/sfu-discourse-lab/SO-CAL/tree/master/Resources/dictionaries/English}}\\
 \cite{taboada2011lexicon}, as retrieved in July 2018.

If a unigram appears in opposite categories for the general and financial sentiment lexicons, we defaulted to the sentiment given by the financial sentiment lexicon. There were 14 instances of terms defined as positive in SO-CAL and negative in Loughran-McDonald: \emph{unpredictably, conviction, correction, force, seriousness, toleration, missteps, overcome, condone, tolerate, exonerate, upset, challenging, unpredictable}. 

We also deleted \emph{question} and \emph{questions} from the negative Loughran-McDonald list since these were abundant in the question-answer portions of earnings calls.

\subsubsection{Hedging}
We used the unigram and ngram hedging dictionaries from
 {\small \url{https://github.com/aproko/hedge_nn}}
 \cite{prokofieva2014hedging}, as retrieved in July 2018.

\subsubsection{Uncertainty, Litigiousness, Modal, Constraining}

We used the word lists from
 {\small \url{https://sraf.nd.edu/textual-analysis/resources/}}\footnote{Archived at {\small \url{https://web.archive.org/web/20181203160914/https://sraf.nd.edu/textual-analysis/resources/}}}
 \cite{loughran2011liability}, as retrieved in July 2018. 

\section{Other modeling experiments}
 
For the prediction task in \S5, in addition to ridge regression and logistic regression, we also experimented with Gaussian kernel ridge regression and support vector machines but found they performed worse or similarly. See Table~\ref{t:results1} for the full results. 

\subsection{Gaussian kernel ridge regression.}
Kernel ridge regression combines ridge regression with the kernel trick and we implement the model with \texttt{sklearn}. We use a Gaussian (RBF) kernel.
To tune hyperparameters, we perform a five-fold cross-validation grid search over the regularization strength, $\alpha$, and the inverse of the radius of influence of samples selected by the model as support vectors, $\gamma$.

\subsection{SVC with RBF kernel.} 
We also train a support vector classifier (SVC) with an RBF kernel and we implement the model with \texttt{sklearn}. 
We tune the hyperparameters ``C" (the penalty parameter of the error term) and gamma (free parameter of the Gaussian radial basis function). The SVM trained on the bag-of-words features ran out of memory, even on a machine with a large amount of RAM.

\setlength{\tabcolsep}{5.5pt}
